\documentclass[conference]{IEEEtran}
\IEEEoverridecommandlockouts
\usepackage{cite}
\usepackage{amsmath,amssymb,amsfonts}
\usepackage{algorithmic}
\usepackage{graphicx}
\usepackage{textcomp}
\usepackage{xcolor}
\def\BibTeX{{\rm B\kern-.05em{\sc i\kern-.025em b}\kern-.08em
    T\kern-.1667em\lower.7ex\hbox{E}\kern-.125emX}}
\begin{document}

\title{Ensemble of radiomics and ConvNeXt for breast cancer diagnosis\\

\thanks{This research was supported by the Secretaría de Ciencia, Humanidades, Tecnología e Innovación (Secihti), with cloud computing resources provided through Microsoft’s AI for Good Lab.}
}

\author{\IEEEauthorblockN{1\textsuperscript{st} Jorge Alberto Garza-Abdala}
\IEEEauthorblockA{\textit{School of Science and Engineering} \\
\textit{Tecnologico de Monterrey}\\
Monterrey, Mexico \\
https://orcid.org/0000-0003-1952-5977}
\and
\IEEEauthorblockN{2\textsuperscript{nd} Gerardo Alejandro Fumagal-González}
\IEEEauthorblockA{\textit{School of Science and Engineering} \\
\textit{Tecnologico de Monterrey}\\
Monterrey, Mexico \\
https://orcid.org/0009-0008-2041-8056}
\and
\IEEEauthorblockN{3\textsuperscript{rd} Beatriz A Bosques-Palomo}
\IEEEauthorblockA{\textit{School of Science and Engineering} \\
\textit{Tecnologico de Monterrey}\\
Monterrey, Mexico \\
a01039875@tec.mx}
\and
\IEEEauthorblockN{4\textsuperscript{th} Mario Alexis Monsivais Molina}
\IEEEauthorblockA{\textit{School of Science and Engineering} \\
\textit{Tecnologico de Monterrey}\\
Monterrey, Mexico \\
a00822468@tec.mx}
\and
\IEEEauthorblockN{5\textsuperscript{th} Daly Avedano}
\IEEEauthorblockA{\textit{School of Medicine and Health Sciences} \\
\textit{Tecnologico de Monterrey}\\
Monterrey, Mexico \\
a00809192@tec.mx}
\and
\IEEEauthorblockN{6\textsuperscript{th} Servando Cardona-Huerta}
\IEEEauthorblockA{\textit{School of Medicine and Health Sciences} \\
\textit{Tecnologico de Monterrey}\\
Monterrey, Mexico \\
servandocardona@tec.mx}
\and
\IEEEauthorblockN{7\textsuperscript{th} José Gerardo Tamez-Pena}
\IEEEauthorblockA{\textit{School of Medicine and Health Sciences} \\
\textit{Tecnologico de Monterrey}\\
Monterrey, Mexico \\
jose.tamezpena@tec.mx}
}

\maketitle

\begin{abstract}
Early diagnosis of breast cancer is crucial for improving survival rates. Radiomics and deep learning (DL) have shown significant potential in assisting radiologists with early cancer detection. This paper aims to critically assess the performance of radiomics, DL, and ensemble techniques in detecting cancer from screening mammograms. Two independent datasets were used: the RSNA 2023 Breast Cancer Detection Challenge (11,913 patients) and a Mexican cohort from the TecSalud dataset (19,400 patients). The ConvNeXtV1-small DL model was trained on the RSNA dataset and validated on the TecSalud dataset, while radiomics models were developed using the TecSalud dataset and validated with a leave-one-year-out approach. The ensemble method consistently combined and calibrated predictions using the same methodology. Results showed that the ensemble approach achieved the highest area under the curve (AUC) of 0.87, compared to 0.83 for ConvNeXtV1-small and 0.80 for radiomics. In conclusion, ensemble methods combining DL and radiomics predictions significantly enhance breast cancer diagnosis from mammograms.
\end{abstract}


\section{Introduction}
Breast cancer is a significant global health issue for women, with 2.3 million cases and 685,000 deaths reported in 2020 \cite{b1}. The disease’s complexity and variability often result in delayed detection, which limits the effectiveness of available treatments \cite{b2}. Early diagnosis is essential for improving outcomes, particularly for women at higher risk, making regular screening vital.

Mammography is the standard screening method; however, its sensitivity decreases in women with dense breast tissue, making cancer diagnosis difficult \cite{b3}. Radiomics addresses this limitation by extracting quantitative features from medical images, enhancing cancer detection, prognosis, and risk assessment \cite{b4, b5, b6}. When combined with machine learning, radiomics can further improve classification accuracy \cite{b7}. Deep learning techniques, such as ConvNeXtV1-small, also contribute to better diagnostic performance. ConvNeXtV1-small aims to match the capabilities of vision Transformers while preserving the simplicity of traditional convolutional networks, making it particularly suitable for breast cancer diagnosis \cite{b8}.

This study proposes that combining radiomics with ConvNeXtV1-small could further enhance diagnostic accuracy. The approach involves developing separate models using radiomics and deep learning and then merging their predictions through an ensemble method.

\section{Dataset description}

This study used two independent datasets. The first dataset was the Radiological Society of North America (RSNA) Screening Mammography Breast Cancer Detection Challenge 2023 dataset which contains 11,913 screening mammograms: 11,665 for negative cancer cases and 248 for positive cancer cases. A detailed explanation of the dataset can be found in \cite{b9}.  

The second dataset, known as the TecSalud dataset, was approved by the local institutional ethics committee (protocol number: P000542-MIRAI-MODIFICADO-CEIC-CR002). It includes 19,031 negative mammograms and 369 positive mammograms (131 BIRADS 4 and 238 BIRADS 5), collected between 2014 and 2019. The median age for the positive cases was 50.26 years, with a standard deviation of 9.93 while for the negative cases was 54.26 and 12.52, respectively. To maintain consistency and avoid bias in model training, only the first mammogram per patient was included in the training process.

Both datasets provide two standard mammographic views for each breast—craniocaudal (CC) and mediolateral oblique (MLO)—for both the right and left sides. They also contain associated patient and image metadata such as age, presence of implants, and BIRADS scores. It is important to notice that no mammograms with implants were used for this study.

\section{Methodology}

\subsection{Radiomics Image Preprocessing}\label{AA}
All image pixel intensities were normalized between 0 and 1, and then the image resolution was set to 0.1 mm per pixel. The left CC and MLO views were mirrored to ensure consistent orientation. Finally, to account for intensity variations across different imaging vendors, a customized histogram matching approach was applied, using the RSNA dataset as the reference. Histogram matching used the intensity distribution of a source image and a reference cumulative distribution functions (CDFs) \cite{b10}. We estimated the reference CDF from 100 random images of the RSNA 2023 challenge. Figure 1 shows a sample of the histogram matching result. 

\begin{figure}[!h]

\begin{minipage}[b]{1.0\linewidth}
  \centering
  \centerline{\includegraphics[width=8.5cm]{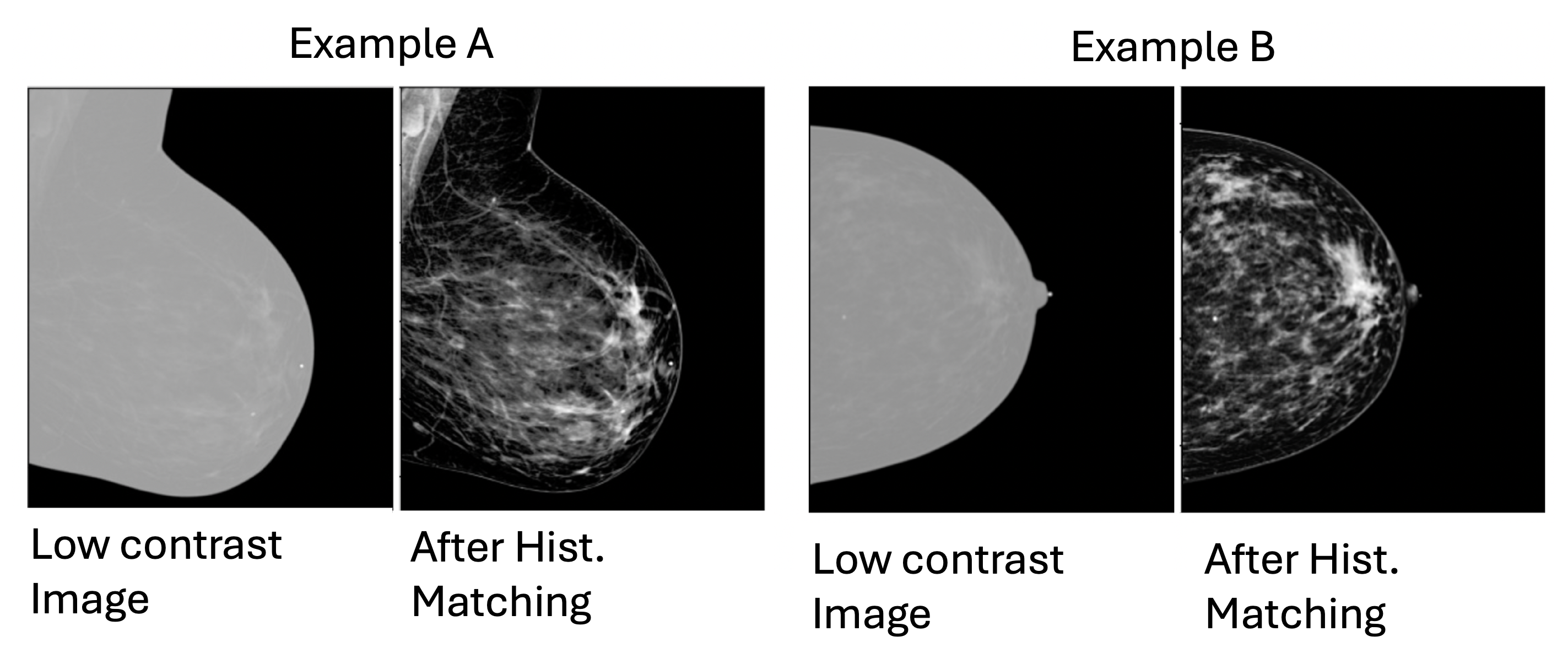}}
  \caption{Sample images from TecSalud dataset before and after applying the preprocessing steps (including histogram matching).}
\end{minipage}
\label{hist_match}
\end{figure}

\subsection{Radiomics Features And Machine Learning}
Radiomics analysis was performed on two regions of interest (ROIs). The automated definition of the ROIs involved three steps: background extraction, breast periphery identification, and ROI selection. Background pixels were identified by a value of zero, while the breast periphery was defined as all breast tissue within 2.0 mm of the background. Within the remaining tissue, two regions were determined using median thresholding to differentiate between dense and non-dense areas for each mammogram view. The dense region provided insights into potential breast cancer tumors, while the non-dense region represented normal fatty tissue characteristics \cite{b11}.
Morphological features such as shape and size were extracted from both dense and non-dense ROIs. Additionally, first-order statistical features (e.g., skewness, kurtosis, entropy) and second-order statistical features based on the gray-level co-occurrence matrix (GLCM) were computed. Wavelet decomposition was applied to the images to capture both coarse and fine details by breaking them down into different frequency components. This process produced four sets of coefficients—attenuation, horizontal, vertical, and diagonal—that highlighted feature strength across various orientations.
For classification, a soft-voting sub-ensemble was developed using multiple machine learning models: K-nearest neighbors (K-NN), Support Vector Machine (SVM), Least Absolute Shrinkage and Selection Operator (LASSO), Bootstrapped Stage-wise Model Selection (BSWiMS), Linear Discriminant Analysis (LDA), Naïve Bayes (NB), and Random Forest (RF). The input to the sub-ensemble included a combination of radiomic features and mammogram acquisition parameters. Given that the TecSalud dataset spans from 2014 to 2019, a leave-one-year-out validation strategy was used for model evaluation.

\subsection{Deep Learning Workflow}
The ConvNeXtV1-small was selected for the deep learning task since it has shown potential for classifying mammograms as seen in \cite{b12}. The model was trained from scratch using the full 2023 RSNA dataset, without any pre-trained weights. To address class imbalance and ensure generalization, model calibration was performed using a leave-one-year-out strategy on the TecSalud dataset. After preprocessing and histogram matching, each mammogram view, along with its corresponding assessment of breast cancer presence, was included in the training process. The model was trained with a learning rate of 0.00015, and a batch size of 192, for 24 epochs. A Cosine-Annealing-LR scheduler was used to adjust the learning rate, while the AdamW optimizer and binary cross-entropy loss function guided the training.

The trained RSNA model was then used to predict the probability of breast cancer for each mammogram view in the TecSalud dataset. To calculate the patient-level probability, the highest value from the four independent evaluations (one for each mammogram view) was selected.

\subsection{Ensemble workflow (Radiomics + ConvNeXtV1-small model)}

Figure 2 illustrates the ensemble model that combines radiomics and DL approaches to predict each patient's cancer probability. The ensemble required calibration of the estimated probabilities to ensure accuracy since the case ratio was different between both datasets. In this study, a leave-one-year-out procedure was applied to calibrate the probabilities from both the DL and radiomics methods. The final per-patient risk of breast cancer was then predicted by averaging the calibrated probabilities

\begin{figure*}[!h]

\begin{minipage}[b]{1.0\linewidth}
  \centering
  \centerline{\includegraphics[width=\textwidth]{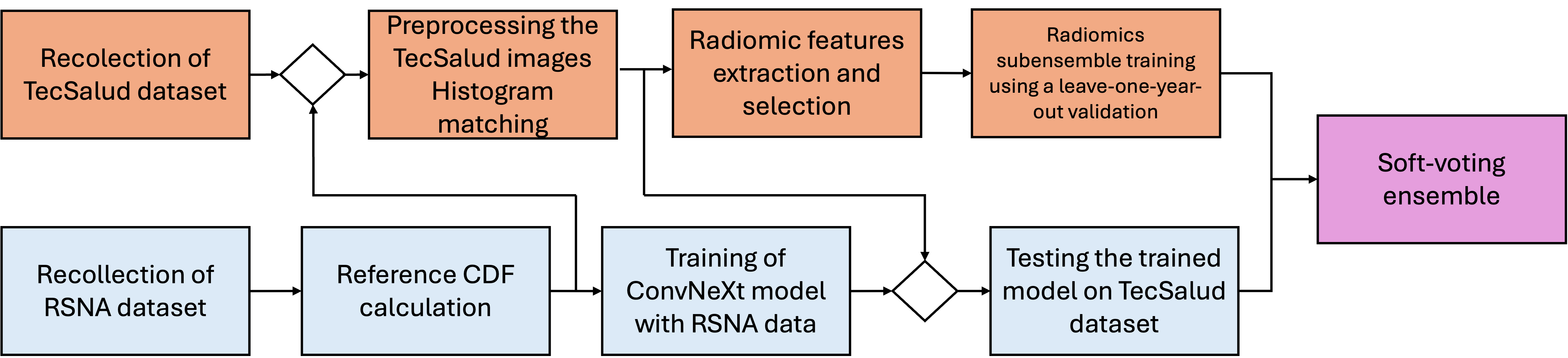}}
  \caption{Workflow of the proposed ensemble. The orange squares represent the radiomics subensemble while the blue squares are for the ConvNeXtV1-small.}
\end{minipage}
\label{Ensemble_wf}
\end{figure*}

\section{Results}
Figure 3 and Table I show the performance of the models using the leave-one-year-out validation strategy. The ensemble model achieved the highest area under the curve (AUC), with a value of 0.878 (95\% CI: 0.859–0.897). This result was better than both the radiomics-only model (AUC = 0.801, 95\% CI: 0.776–0.825) and the ConvNeXtV1-small model (AUC = 0.830, 95\% CI: 0.815–0.863). In addition, the ensemble model reached a true positive rate (TPR) of 0.778 and a true negative rate (TNR) of 0.831. The overall accuracy was 0.830, with an F1-score of 0.148 and a balanced error rate (BER) of 0.196.

Table II provides a comparison of the AUC values for the radiomics sub-ensemble, the ConvNeXtV1-small model, and the proposed ensemble method.

\begin{table}[h]
    \centering
    \caption{Confusion matrix for the ensemble}
    \label{tab:conf_matrix}  
    \resizebox{\linewidth}{!}{
    \begin{tabular}{|l|c|c|c|}
        \hline
        & \textbf{Outcome +} & \textbf{Outcome -} & \textbf{Total} \\
        \hline
        \textbf{Test +} &  287 & 3,219  & 3,506  \\
        \hline
        \textbf{Test -} &  82  & 15,812 & 15,894 \\
        \hline
        \textbf{Total}  &  369 & 19,031 & 19,400 \\
        \hline
    \end{tabular}
    }
\end{table}

\begin{table}[h]
    \centering
    \caption{AUC and Confidence Intervals for the radiomics subensemble, the ConvNeXtV1-smallV-1, and the ensemble of both.}
    \resizebox{0.8\linewidth}{!}{
        \begin{tabular}{|l|c|c|c|}
        \hline
        \textbf{Model} & \textbf{AUC (95\% CI)} \\
        \hline
        RAD & 0.801 (0.776-0.825) \\
        \hline
        DL  & 0.839 (0.815-0.863) \\
        \hline
        ENS & 0.878 (0.859-0.897) \\
    \hline    
    \end{tabular}
    }
\end{table}

\begin{figure}[htb]
\begin{minipage}[b]{1.0\linewidth}
  \centering
  \centerline{\includegraphics[width=\textwidth]{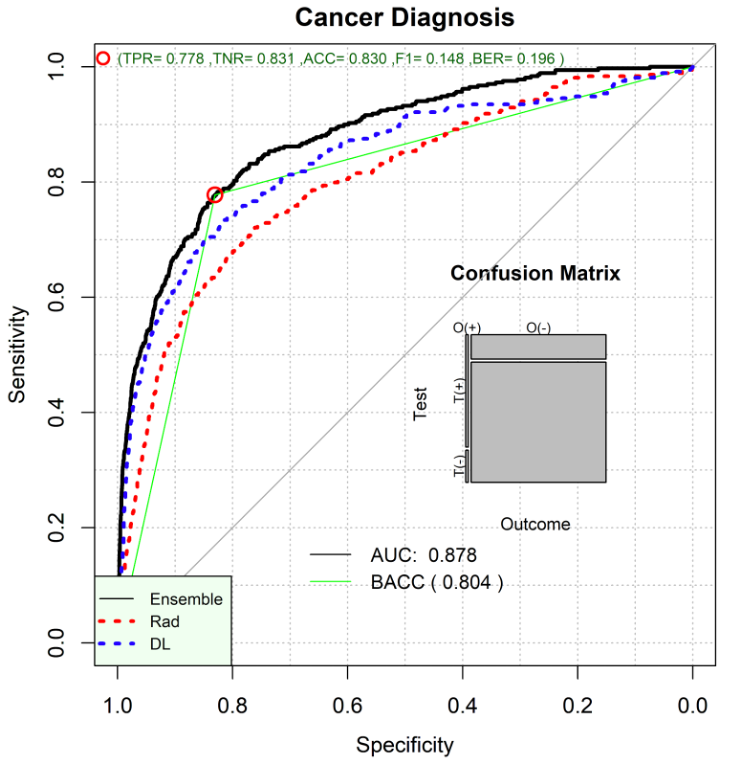}}
  \caption{AUC ROC for the ensemble. The dotted red line represents the radiomics sub ensemble, while the dotted blue line is the ConNeXt.}
\end{minipage}

\end{figure}

\vspace{0.25cm}

\section{Discussion}
This study presented a radiomics–DL ensemble for breast cancer diagnosis, which significantly improved the performance of the individual models. The ConvNeXtV1-small was trained from scratch using the RSNA 2023 dataset, while the radiomics model was trained exclusively on the TecSalud dataset. Validation followed a leave-one-year-out strategy, ensuring independence between training and test sets and simulating a real-world clinical workflow. The ensemble achieved a TPR of 0.778, TNR of 0.831, accuracy of 0.83, F1-score of 0.148, BER of 0.196, and an AUC of 0.878 (95\% CI: 0.859–0.897), outperforming the individual models (AUC = 0.80 for radiomics-only and 0.83 for ConvNeXtV1-small). AUC was prioritized as the main comparison metric due to its robustness to class imbalance. These findings support the value of integrating handcrafted and learned features for improved diagnostic accuracy.

Previous research has reported similar benefits from combining radiomics and DL. For example, Beque et al. combined a ResNet101 for segmentation and classification with an XGBoost model trained on 1,320 handcrafted features. Their final predictions averaged the outputs of both models, reaching an AUC of 0.95 with DL-based segmentations. Although validated externally, the sample size (279 patients) may be insufficient for broader generalization \cite{b13}. Similarly, Lu et al. developed a multimodal ensemble using ultrasound, mammography, and MRI, integrating 108 radiomic features and a pretrained ResNet-50. Their stacked model achieved an AUC of 0.997 and perfect specificity but required multiple imaging types and was limited to 322 patients \cite{b14}.

Chouhan et al. used handcrafted descriptors and CNN features from ROIs on the DDSM dataset, proposing an emotional learning–inspired ensemble combining k-NN and SVM. Their model achieved an AUC of 0.865 but was constrained by its reliance on ROIs rather than full mammograms \cite{b15, b16}. Haq et al. applied a voting ensemble of a DCNN, SVM, and random forest, trained on MIAS and validated on BCDR \cite{b17, b18}. Despite achieving strong performance, their approach used only DL features and ROI-based inputs \cite{b19}.

Compared to these studies, our work stands out by using two larger datasets (RSNA and TecSalud), focusing only on mammography, and applying a simpler model structure: one deep learning network and a sub-ensemble of classic machine learning models. This makes the method less demanding in terms of computational resources and potentially easier to apply in clinical settings.

\subsection{Limitations and future work}
The main limitation of the presented study is the use of future data for the prediction of previous year mammograms. Future work will use strict train and external validation schemes to train the ensemble model allowing for a realistic evaluation of the methodology on external datasets. Furthermore, the contribution of each radiomics feature will also be studied

\section{Conclusion}

This study confirmed the hypothesis that combining radiomics and deep neural networks in an ensemble model could improve brest cancer diagnostic performance from screening mammograms. The proposed ensemble achieved a higher AUC compared to its sub-models, demonstrating its effectiveness. The use of multiple datasets for external validation was also essential, as it enhances the model's generalizability and potential for clinical application.

\section*{Compliance with ethical standards}
\label{sec:ethics}

The study was approved by the ethical committee Comite
de Ética en Investigación de la Escuela de Medicina del Instituto Tecnologico y de Estudios Superiores de Monterrey. Informed consent was obtained from all the participants and/or their legal guardians. All experiments were performed in accordance with relevant guidelines and regulations of the
Declaration of Helsinki.

\section*{Acknowledgment}
\label{sec:acknowledgments}

This research was supported by the Secretaría de Ciencia, Humanidades, Tecnología e Innovación (Secihti), with cloud computing resources provided through Microsoft’s AI for Good Lab.

\end{document}